# Collective Anomaly Detection based on Long Short Term Memory Recurrent Neural Network

Loïc Bontemps, Van Loi Cao, James McDermott, and Nhien-An Le-Khac

University College Dublin, Dublin, Ireland
loic.bontemps@ucdconnect.ie,loi.cao@ucdconnect.ie,james.mcdermott2@ucd.ie,an.lekhac@ucd.ie

**Abstract.** Intrusion detection for computer network systems becomes one of the most critical tasks for network administrators today. It has an important role for organizations, governments and our society due to its valuable resources on computer networks. Traditional misuse detection strategies are unable to detect new and unknown intrusion. Besides, anomaly detection in network security is aim to distinguish between illegal or malicious events and normal behavior of network systems. Anomaly detection can be considered as a classification problem where it builds models of normal network behavior, which it uses to detect new patterns that significantly deviate from the model. Most of the current research on anomaly detection is based on the learning of normally and anomaly behaviors. They do not take into account the previous, recent events to detect the new incoming one. In this paper, we propose a real time collective anomaly detection model based on neural network learning and feature operating. Normally a Long Short-Term Memory Recurrent Neural Network (LSTM RNN) is trained only on normal data and it is capable of predicting several time steps ahead of an input. In our approach, a LSTM RNN is trained with normal time series data before performing a live prediction for each time step. Instead of considering each time step separately, the observation of prediction errors from a certain number of time steps is now proposed as a new idea for detecting collective anomalies. The prediction errors from a number of the latest time steps above a threshold will indicate a collective anomaly. The model is built on a time series version of the KDD 1999 dataset. The experiments demonstrate that it is possible to offer reliable and efficient for collective anomaly detection.

## 1 Introduction

Network anomaly detection refers to the problem of detecting illegal or malicious activities or events from normal connections or expected behavior of network systems [4, 5]. It has become one of the most concerned subjects in network security domain due to the fact that organizations or governments are now seeking for good solutions to protect their valuable resources on computer networks from unauthorized and illegal accesses, network attacks or malware. Over the last three decades, machine learning techniques are known as a common approach for developing network anomaly detection models [3, 4]. Network anomaly detection is usually posed as a type of classification problem: given a dataset representing normal and anomalous examples, the goal is to build a learning classifier which is capable of signaling when a new anomalous data sample is encountered [5].

However, most of the existing approaches consider an anomaly as a single point: cases when they occur "individually" and "separately" [6, 7, 16]. In such approaches, anomaly detection models do not have the ability to represent the information from previous data or events for evaluating a current point. In network security, some kinds of attacks, *Denial of Service (DoS)*, usually occur for a long period of time (several minutes) [10], and are often represented by a set of single points. An attack will be indicated only if a set of single points are considered as attack. In order to detect this kinds of attacks, anomaly detection models should be capable of remembering the information from a number of previous events, and representing the relationship between them and current event. To avoid important mistakes, one must always consider every outcome: in this sense a highly anomalous value may still be linked to a perfectly normal condition, and conversely. In this work, we aim to build an anomaly detection model for this kinds of attacks (known as collective anomaly mentioned in [5]).

Collective anomaly is the term to refer to a collection of related anomalous data instances with respect to the whole dataset [5]. The single data points in a collective anomaly may not be considered as anomalies by themselves, but the occurrence of these single points together indicates an anomaly. Long Short Term Memory Recurrent Neural Network (LSTM RNN) is known as one of powerful techniques to represent the relationship between current event and previous events, and handles time series problems [12, 14]. Thus, it is employed to develop anomaly detection model in this paper.

In this paper, we will propose a collective anomaly detection model by using the predictive power of LSTM RNN [8]. Firstly, LSTM RNN is applied as a time series anomaly detection model. The prediction of a current event will depend on both the current event and its previous events. Secondly, the model will be adapted to detect collective anomaly by proposing a circular array. The circular array contains the prediction errors from a certain number of latest time steps. If the prediction errors in the circular array are higher than predeterminer threshold and last for a certain time steps, it will indicate a collective anomaly. More details will be described in Section 4.

The rest of the paper is organized as follows. We briefly review some work related to anomaly detection and LSTM RNN. In Section 3, we give a short introduction to LSTM RNN. This is followed by a section proposing the collective anomaly detection model using LSTM RNN. Experiments, Results and Discussion are presented in Section 5 and Section 6 respectively. The paper concludes with highlights and future directions.

## 2   Related Work

When considering a time series dataset, point anomalies are often directly linked to the value of the considered sample. However, attempting real time collective anomaly detection implies always being aware of previous samples, and more precisely their behavior. This means that every time step should include an evaluation of the current value combined with the evolution of precedent information. In this section, we briefly describe work related applying LSTM RNN to time series and collective anomaly detection problems [12, 14, 15].

Olsson et al. [15] proposed an unsupervised approach for detecting collective anomaly. In order to detect a group of the anomalous examples, the "anomalous score" of the group of data points was probabilistically aggregated from the contribution of each individual examples. Obtaining the collective anomalous was processed under unsupervised manner, thus it is suitable for both unsupervised and supervised learning anomaly techniques to scoring individual anomalies. The model was evaluated on an artificial dataset and two industrial datasets, detecting anomalies in moving cranes and anomalies in fuel consumption.

In [12], Malhotra et al. applied LSTM network for addressing the problem of time series anomaly detection. The stacked LSTM network trained on only normal data was used to predict over a number of time steps. They assumed that the resulting prediction errors has a Gaussian distribution, which was used to assess the likelihood of anomaly behavior. Their model was demonstrated performing well on four datasets.

Marchi et al. [14, 13] presented a novel approach by combining non-linear predictive denoising autoencoders (DA) with LSTM for identifying abnormal acoustic signals. Firstly, LSTM Recurrent DA was employed to predict auditory spectral features of the next short-term frame from its previous frames. The network trained on normal acoustic recorders tends to behave well on normal data, and yields small reconstruction errors whereas the reconstruction errors from abnormal acoustic signals are high. The reconstruction errors of the autoencoder was used as "anomaly score", the reconstruction error above a pre-determine threshold indicates an novel acoustic event. The model was trained on a public dataset containing in-home sound events, and evaluated on a dataset including new more anomaly events. The results demonstrated that their model performed significantly better than exsiting methods.

The idea is also used in a practical acoustic example [14, 13], where LSTM RNNs are used to predict short-term frames. The core idea of this paper is to combine the previous methods, to adapt Long Short-Term Memory to collective anomaly detection. By labelling testing LSTM RNN outputs at every time step with a standardized error value, we shall propose an algorithm to detect collective anomalies. This will prove very useful in our example : First, we will train normal data on an LSTM RNN in order to estimate the behaviour of a normal day of traffic. Then, we will use a classifier inspired by [15] to rate the level of anomaly of each time sample. We will apply this method to a network security problem (KDD 1999 cup), aiming to raise an alarm in the case of DoS Neptune attacks.

# 3 Preliminaries

In this section, we briefly describe a specific type of Recurrent Neural Network: Long Short Term Memory. The structure was proposed by Hochreiter et al. [8] in 1997, and has already proven as a powerful technique for addressing the problem of time series prediction.

The difference initiated by LSTM regarding other types of RNN resides in its "smart" nodes presented in Fig. 1. Each of these cells contains three gates, input gate, forget gate and output gate, which decide how to react to an input. Depending on the strength of the information each node receives, it will decide to block it or pass it on. The information is also filtered with the set of weights associated with the cells when it is transferred through these cells.

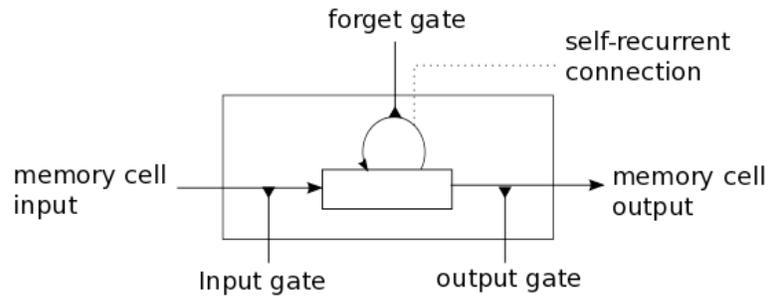

**Fig. 1.** LSTM RNN Cell, figure reproduced from [1]

The LSTM node structure enables a phenomenon called backpropagation through time. By calculating for each hidden layer the partial derivatives of the output, weight and input values, the system can move backwards to trace the evolving error between real output and predicted output. Afterwards, the network uses the derivative of this evolution to adapt its weights and decrease prediction error. This learning method is named Gradient Descent.

As mentioned before, Long Short-Term Memory has the power to incorporate a behaviour into a network by training it with normal data. The system becomes representative of the variations of the data. In other words, a prediction is made focusing on two features: the value of a sample and its position at specific time. This means that two input samples at different times may have the same value, but their outputs will very probably differ. It is because a LSTM RNN is able to use Back Propagation through time to consider the past of a sample, and therefore understand its context better.

## 4 Proposed Approach

In this section, we are going to describe a new approach to address the problem of collective anomaly detection. Firstly, we show the LSTM RNNs ability to impregnate itself with the behaviour of a training set, and in this stage it acts like a time series anomaly detection model. We will then adapt it for collective anomaly detection by introducing terms that measure its prediction errors in a period of time steps. Finally, we shall describe how to seek a collective anomaly by combining a LSTM RNN with a circular array method.

### 4.1 LSTM RNN as a predictive vector

The first step inspires the idea presented in [12]: when trained correctly, LSTM RNNs have the ability to impregnate themselves with the behavior of a training set. Intuitively meaning that when given a certain input samples, they have the ability to remember the context of the value of the samples, and to predict a coherent output in agreement with the context of the sample. In our work, we will use a simple LSTM RNN, in opposition to stacked LSTM in [12]. This does not change the core principle of the method: when given sufficient training, a LSTM RNN adapts its weights, which become representatives of the training data.

### 4.2 Definitions

In order to adapt a LSTM RNN for time series data to detect collective anomalies, we introduce terms to measure prediction errors at each time step or in a period of time steps. These terms are defined as below.

- **Relative Error (RE):** the Relative Error between two real values $x$ and $y$ is given by Eq. 1:

$$RE\ (x,\ y) = \frac{|x - y|}{x} \quad (1)$$

- **Relative Error Threshold (RET):** Relative Error value above a predetermined threshold indicates an anomaly. This threshold, $RET$, is determined by using labeled normal and attacks from a validation set.
- **Minimum Attack Time (MAT):** The minimum amount of recent time steps that is used to define a collective attack.
- **Danger Coefficient (DC):** The density of anomalous points within the last $MAT$ time steps. Let $N$ be the number of anomalous points over the last $MAT$ time steps, $DC$ is defined as in Eq. 2.

$$DC = \frac{N}{MAT} \quad (2)$$

NB: $0 < DC < 1$

– **The Averaged Relative Error (ARE):** The Average Relative Error over a $MAT$ is given by Eq. 3:

$$ARE = \sum_{i=1}^{MAT} RE_i \tag{3}$$

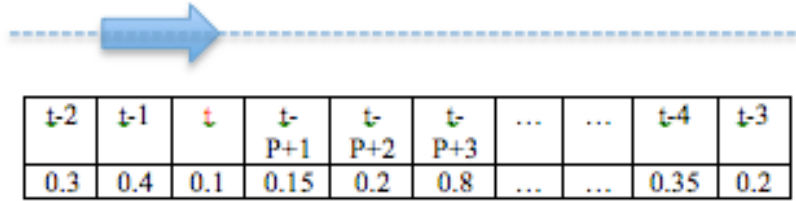

**Fig. 2.** Circular Array for Collective Anomaly Detection model, $MAT = P$.

The values of two terms, *Danger Coefficient* and *Average Relative Error*, are the keys factors that will help the model to decide whether a set of inputs within a number of the latest time steps is collective anomaly or not as described in 4.3. These values will be estimated by using validation set.

### 4.3 Degree of error evaluation

At each time step, the sample predicted by the LSTM RNN is compared with the real future sample. This comparison is computed as a $RE$ value. In this sense, a "Relative Error time series" is built in live. Based on the observation of a validation set, we can initialize value for the $RET$.

At this stage, our system is theoretically capable of detecting point anomalies at each time step. In order to adapt the model from a individually anomaly to collective anomaly, we must consider simultaneously an ensemble of points. To do this, we propose a circular array containing the $MAT$ latest error values to represent the level of anomaly of the latest time steps as shown in Fig. 2. By analyzing the circular array at every time step, we evaluate the possibility of facing a collective anomaly. A collective anomaly will be identified if both *Danger Coefficient* and *Average Relative Error* are higher than predefined thresholds, $\alpha$ and $\beta$, respectively. It is formulated in Eq. 4 and 5 as below.

$$DC > \alpha \tag{4}$$

$$ARE > \beta \tag{5}$$

where $\alpha$ and $\beta$ will be estimated by using the validation set.

Once training is terminated, the thresholds and parameters of the LSTM network will be adjusted correctly (in order to obtain a satisfactory error decrease). The model should be able to determine with good efficiency whether a set of points represents a collective anomaly.

## 5 Experiments

### 5.1 Datasets

In order to demonstrate the efficient performance of the proposed model, we choose a dataset related to network security domain, the KDD 1999 dataset [2, 9], for our experiments. The dataset in tcpdump format was collected from a simulated military-like environment over a period of 5 weeks. There are four main groups of attacks in the dataset, but we restrict our experiments on a specific attack, *Neptune*, in Denial-of-Service (DoS). The dataset is also converted into time series version before feeding into the model. More details about how to obtain a time series version from the original dataset, and how to choose training, validation and testing sets are presented in the following paragraphs.

The first crucial step is to build a conveniently usable time series dataset out of the tcpdump data, and selecting the features we wish to use. We use terminal commands and a python program to convert the original tcpdump records in the KDD 1999 dataset into a time dependant function. This method is a development of the proposed transformation in [11] that acts directly on the tcpdump to obtain real time statistics of the data. Our scheme follows this step by step transition as described below:

$$\text{tcpdump} \Rightarrow \text{pcap} \Rightarrow \text{csv}$$

Each day of records can be time-filtered and input into a new *.pcap* file. This also has the advantage of giving a first approach on visualizing the data by using Wireshark functionalities (IO graphs and filters). Once this is done, the *tshark* command is adapted to select and transfer the relevant information from the records into a *.csv* file. We may note that doing this is a first step towards faster computation and better system efficiency, since all irrelevant pcap columns can be ignored. There are two major steps for the conversion processing.

1. Store the information of a *.tcpdump* file into a newly generated *.pcap* file. From the terminal, we use the *editcap* command:

    ```
    editcap -A '1999-03-11 08:00:00' -B'1999-03-11 18:00:00'
    Thursday2outside.tcpdump Thursday2.pcap
    ```

2. Convert from *.pcap* file into *.csv* file by *tshark* command. From the terminal again, type the command below:

    ```
    tshark -r Thursday2.pcap -T fields -e frame.number -e frame.len
    -e frame.time -e ip.proto -E header=y -E separator=, -E quote=d
    -E occurrence=f -i netstat -f tcp[13]==12  > Thursday2.csv
    ```

*tshark* is a simple but powerful command, enabling the selection of columns of interest in a *.pcap* file, and their output in a newly generated *.csv*. Once the data is in the *.csv* format, python code can be implemented from the XX library to store it and use with our classifier.

Processing the tcpdump with this method enables quick and easy manipulation of the data. For example, Neptune and Smurf are both DoS attacks caracterized by a high flow of specific packets in networks (eg. SYN_ACK and ICMP echo replies). By using this simple fact, the needed records can be filtered and counted at every time step. If we aim to detect Neptune attack, the *thark* command can be implemented with the $-i$ netstat $-f$ $tcp[13] == 2$ filter, so only SYN_ACK packets from servers are counted. We observe in the case of KDD 1999 that a Neptune attack can be sought by looking for an anomalously high number of these packets.

The KDD1999 time series is composed of a two-weeks training set $n_1$ (weeks 1 & 3, normal data), one week of validation set $v_1$ (week 2, both labeled normal and anomaly data), and a two-weeks of testing set $t_1$ (weeks 4 & 5). The protocol will be the following: training the network with $n_1$, using $v_1$ to determine our error threshold(s), and evaluating the proposed model on $t_1$.

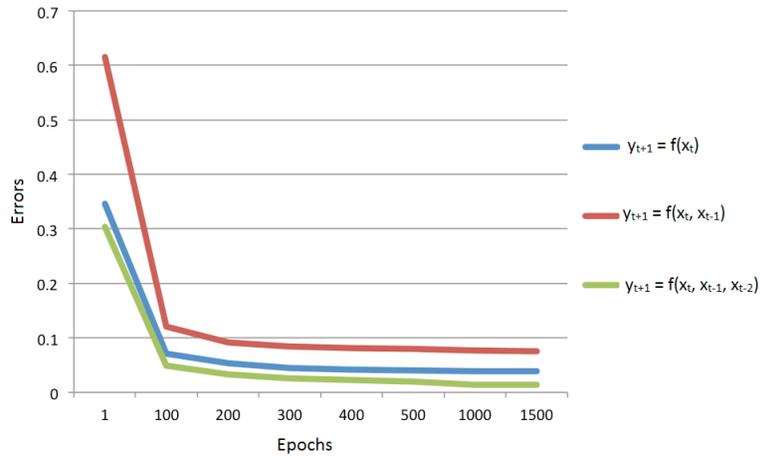

**Fig. 3.** The training errors from the model with one, two and three inputs.

### 5.2 Experimental Settings

In this work, we conduct two experiments, one preliminary experiment and one main experiment. The preliminary experiment is aim to estimate the parameters for the models and set its thresholds by using the validation set whereas the main experiment is to evaluate the proposed model.

**Preliminary Experiment:**

This experiment is aim to select the best parameters of our LSTM RNN model with respect to minimize its prediction error, and determine the thresholds, $\alpha$ and $\beta$. Firstly, we determine how many the previous time steps should be used for predicting the current event. The hyper-parameters of LSTM RNN, hidden size and learning rate, is then estimated. Finally, the two thresholds, $\alpha$ and $\beta$, will be estimated with respect to a good classification performance of the model on the validation set.

In order to optimize the proposed model for the main experiment, we proceed to a preliminary test to measure the influence of the number of inputs on the prediction error of LSTM. We first focus on how many inputs will influence the prediction of an LSTM [12]. More specifically, they show that a sample $x_t$ input at time $t$ will be predicted with reasonable accuracy $y_{t+1}$. We emit the hypothesis that inserting more values in our system may help decrease prediction errors, but it leads to more time consuming. Thus, we investigate the relationship between the prediction value $y_{t+1}$ to three sets of the previous input examples $(x_t)$, $(x_t, x_{t-1})$, $(x_t, x_{t-1}, x_{t-2})$. They are formulated in equations 6, 7 and 8 below:

$$y_{t+1} = f(x_t) \tag{6}$$

$$y_{t+1} = f(x_t, x_{t-1}) \tag{7}$$

$$y_{t+1} = f(x_t, x_{t-1}, x_{t-2}) \tag{8}$$

where $x_t$, $x_{t-1}$ and $x_{t-2}$ are the input samples at times $t$, $t-1$ and $t-2$ respectively, and $y_{t+1}$ is the predicted value for the input $x_t$.

The number of hidden nodes and the learning rate are the final two parameters that can highly influence the performance of a LSTM RNN. On the one hand, the strength of a LSTM RNN resides in its hidden layer. Each synapse of a network is weighted differently, and can be considered as a unique interpretation of the input data. Each node of the hidden layer is storage space for these interpretations. Theoretically, the higher number of hidden nodes, the more information the network can contain. This also means more computation, and may lead to over-fitting.

Using the LSTM RNN error evolution curve empirically, we concluded that the optimum number of nodes in our hidden layer to obtain good memorization is approximatively *23*. The learning rate is other factor directly linked to the speed at which a LSTM RNN can improve its predictions. For a time step $t$ during training, the synapse weights of our neural network are updated. The learning rate defines how much we wish a weight to be modified at each instant. In our experiment, we choose learning rate equal to *0.01* that gives us a convenient error curve.

Finally, the classifier that is trained on ten days of normal data is used to determine $\alpha$ and $\beta$. We observe the reaction of the system on labeled Neptune attacks from the validation set, and set the thresholds. The values of these thresholds is shown in Section 6.

**Main Experiment:**

Our task is to use the potential speed and accuracy of LSTM RNN to detect a disproportionate durable change in a time series. Once the preliminary experiment is complete, we choose the most performant LSTM RNN, and train it with the normal training set $n_1$. The classifier is then evaluated on testing set $t_1$ containing both normal and attack data to investigate how efficiently our proposed classifier performs.

## 6  Results and Discussion

This section presents our experimental results. First, the preliminary experiment evaluates two factors: computation cost and LSTM prediction error when using one input, two inputs and three inputs respectively. Then, the general performance in terms of classification accuracy is measured.

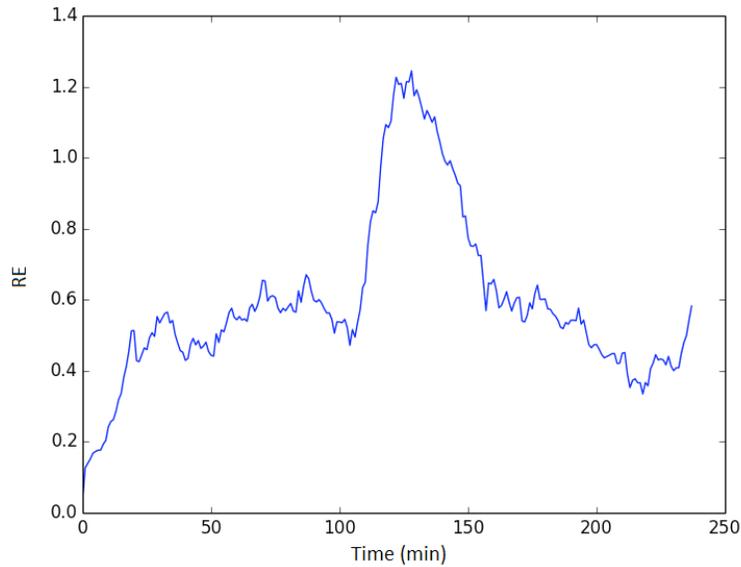

**Fig. 4.** The prediction error from the model with three inputs (1500 Epochs).

The Table 1 illustrates that the model with three inputs had less computational time than those with one or two inputs. Moreover, the Fig. 3 shows that the model with three inputs achieve a lower training error in comparison to two others. Thus, we use the model with three inputs for our main experiment.

**Table 1.** Computational time recording

| Number of inputs | Computational time (seconds) |
|---|---|
| 1 | 645.455811 |
| 2 | 652.4613178 |
| 3 | 642.7698278 |

The results from the main experiment are shown in Table 2. The experiment is done with $MAT = 12$, and $\alpha = 0.66$, and we also report the results on four values of $\beta$, $\beta = 0.69, 0.66, 0.62$ and $0.52$. We observe that it is possible to obtain 100% collective anomaly detection rate, but this infers triggering a high amount of false alarms. Conversely, it is possible to avoid false alarms, but less correct alarms will be detected. Ultimately, detecting more real attacks results in triggering more false alarms as shown in Table 2.

**Table 2.** Circular array detection efficiency

| Threshold $\beta$ | Percentage of correct alarms triggered | Number of false alarms triggered |
|---|---|---|
| 0.69 | 86% | 0 |
| 0.66 | 94% | 2 |
| 0.62 | 98% | 16 |
| 0.52 | 100% | 63 |

## 7 Conclusion and Further work

In this paper, we have proposed a model for collective anomaly detection based on Long Short-Term Memory Recurrent Neural Network. We have motivated this method through investigating LSTM RNN in the problem of time series, and adapted it to detect collective anomaly by proposing the measurements in 4.2. We investigated the hyper-parameters, the suitable number of inputs and some thresholds by using the validation set.

The proposed model is evaluated by using the time series version of the KDD 1999 dataset. The results suggest that proposed model is efficiently capable of detecting collective anomalies the dataset. However, they must be used with caution. The training data fed into a network must be organized in a coherent manner to guarantee the stability of the system. In future work, we will focus on how to improve the classification accuracy of the model. We also observed that implementing variations in a LSTM RNNs number of inputs might trigger different output reactions.

## 8 Acknowledgements

## References

1. LSTM networks for sentiment analysis. In: LSTM networks for sentiment analysis deeplearning 0.1 documentation, http://deeplearning.net/tutorial/lstm.html#lstm, [Accessed 25 Jun 2016]
2. DARPA intrusion detection evaluation. (n.d.). (Retrieved June 30, 2016), http://www.ll.mit.edu/ideval/data/1999data.html
3. Ahmed, M., Mahmood, A.N., Hu, J.: A survey of network anomaly detection techniques. Journal of Network and Computer Applications 60, 19–31 (2016)
4. Bhattacharyya, D.K., Kalita, J.K.: Network anomaly detection: A machine learning perspective. CRC Press (2013)
5. Chandola, V., Banerjee, A., Kumar, V.: Anomaly detection: A survey. ACM computing surveys (CSUR) 41(3), 15 (2009)
6. Chmielewski, A., Wierzchon, S.T.: V-detector algorithm with tree-based structures. In: Proc. of the International Multiconference on Computer Science and Information Technology, Wisła (Poland). pp. 9–14. Citeseer (2006)
7. Hawkins, S., He, H., Williams, G., Baxter, R.: Outlier detection using replicator neural networks. In: International Conference on Data Warehousing and Knowledge Discovery. pp. 170–180. Springer (2002)
8. Hochreiter, S., Schmidhuber, J.: Long short-term memory. Neural computation 9(8), 1735–1780 (1997)
9. KDD Cup Dataset (1999), available at the following website http://kdd.ics.uci.edu/databases/kddcup99/kddcup99.html
10. Lee, W., Stolfo, S.J.: A framework for constructing features and models for intrusion detection systems. ACM transactions on Information and system security (TiSSEC) 3(4), 227–261 (2000)
11. Lu, W., Ghorbani, A.A.: Network anomaly detection based on wavelet analysis. EURASIP Journal on Advances in Signal Processing 2009, 4 (2009)
12. Malhotra, P., Vig, L., Shroff, G., Agarwal, P.: Long short term memory networks for anomaly detection in time series. In: Proceedings. p. 89. Presses universitaires de Louvain (2015)
13. Marchi, E., Vesperini, F., Eyben, F., Squartini, S., Schuller, B.: A novel approach for automatic acoustic novelty detection using a denoising autoencoder with bidirectional lstm neural networks. In: 2015 IEEE International Conference on Acoustics, Speech and Signal Processing (ICASSP). pp. 1996–2000. IEEE (2015)
14. Marchi, E., Vesperini, F., Weninger, F., Eyben, F., Squartini, S., Schuller, B.: Non-linear prediction with lstm recurrent neural networks for acoustic novelty detection. In: 2015 International Joint Conference on Neural Networks (IJCNN). pp. 1–7. IEEE (2015)
15. Olsson, T., Holst, A.: A probabilistic approach to aggregating anomalies for unsupervised anomaly detection with industrial applications. In: FLAIRS Conference. pp. 434–439 (2015)
16. Salama, M.A., Eid, H.F., Ramadan, R.A., Darwish, A., Hassanien, A.E.: Hybrid intelligent intrusion detection scheme. In: Soft computing in industrial applications, pp. 293–303. Springer (2011)